**Discrimination of Radiologists Utilizing Eye-Tracking Technology and Machine Learning: A Case Study**

Stanford Martinez[1], Carolina Ramirez-Tamayo[1], Syed Hasib Akhter Faruqui[2], Kal L. Clark[3], Adel Alaeddini*,[1], Nicholas Czarnek[6], Aarushi Aggarwal[4], Sahra Emamzadeh[3], Jeffrey R. Mock[5], Edward J. Golob[5]

[1] Department of Mechanical Engineering, The University of Texas at San Antonio, Texas, United States.

[2] Department of Radiology, Northwestern University, Chicago, Illinois.

[3] Department of Radiology, University of Texas Health Science Center at San Antonio, Texas, United States.

[4] Long School of Medicine, University of Texas Health Science Center at San Antonio, Texas, United States.

[5] Department of Psychology, The University of Texas at San Antonio, Texas, United States.

[6] Independent Researcher, Anaheim, California, United States.

**Abstract**

Perception-related errors comprise most diagnostic mistakes in radiology. To mitigate this problem, radiologists employ personalized and high-dimensional visual search strategies, otherwise known as search patterns. Qualitative descriptions of these search patterns, which involve the physician verbalizing or annotating the order he/she analyzes the image, can be unreliable due to discrepancies in what is reported versus the actual visual patterns. This discrepancy can interfere with quality improvement interventions and negatively impact patient care. This study presents a novel discretized feature encoding based on spatiotemporal binning of fixation data for efficient geometric alignment and temporal ordering of eye movement when reading chest X-rays. The encoded features of the eye-fixation data are employed by machine learning classifiers to discriminate between faculty and trainee radiologists. We include a clinical trial case study utilizing the Area Under the Curve (AUC), Accuracy, F1, Sensitivity, and Specificity metrics for class separability to evaluate the discriminability between the two subjects in regard to their level of experience. We then compare the classification performance to state-of-the-art methodologies. A repeatability experiment using a separate dataset, experimental protocol, and eye tracker was also performed using eight subjects to evaluate the robustness of the proposed approach. The numerical results from both experiments demonstrate that classifiers employing the proposed feature encoding methods outperform the current state-of-the-art in differentiating between radiologists in terms of experience level. This signifies the potential impact of the proposed method for identifying radiologists' level of expertise and those who would benefit from additional training.

1. **Problem Description**

Lung cancer is the leading cause of cancer death, claiming 139,000 American lives yearly (*An Update on Cancer Deaths in the United States*, 2022). To mitigate its impact, the United States Preventative Task Force recommends annual radiological screening in at-risk individuals (Jones, 2014). Radiologists identify suspicious lung lesions (nodules) from patient chest images and recommend further management, including biopsy, continued surveillance, or further workup. Radiological surveillance reduces population mortality from lung cancer, but it is estimated that radiologists will make errors on 33% of abnormal chest exams, eliminating the chance for patients to start lifesaving treatment (Waite et al., 2019). The predominant source of these errors is not deficient medical knowledge. Rather, errors primarily stem from the methods radiologists use to visually inspect the image, referred to as "perceptual errors" (Bruno et al., 2015). Kundel et al. (Kundel, 1989) investigated the effects of perceptual errors in radiology and concluded that decisions and outcomes improve when radiologists' experiences are enhanced.

Radiologists and radiology educators understand the stakes associated with missed diagnoses due to perceptual errors but have limited tools to combat these errors. Classical educational texts include general concepts, e.g., "...scan the areas of least interest first, working toward the more important areas" ("Felson's Principles of Chest Roentgenology," 2000), which, unfortunately, are inadequate to improve radiologist performance meaningfully.

Eye-tracking technology has been previously proposed as a tool to evaluate radiologist perception. Eye trackers are powerful because they provide high (>30 Hz) temporal and spatial resolution (approximately 1 degree of error). With the aid of eye-tracking, quantitative analyses can be performed to understand the cognitive and perceptual processes better. Eye-tracking technology has previously proven relevant in evaluating decision making processes (Tanoubi et al., 2021), attention interruption (Wu & Wolfe, 2019), skill level determination (Kelly et al., 2016), and impact of search pattern education (Wolfe et al., 2022).

In 2017, Van der Gijp performed a systematic literature review outlining the current state of science concerning visual perception in radiology (van der Gijp et al., 2017). A key tenet is the global-focal search model (Drew et al., 2013; Kundel & Nodine, 1975; Swensson, 1980), which can be summarized as the generation of an initial, fast global impression followed by a more detailed focal search. Eye-tracking technology allows these principles to be tested and potentially optimized to evaluate all clinically relevant portions of the exam in greater detail. Of the 22 relevant articles Van der Gijp reviewed, a consensus "traditional" feature set consisting of 5 features that could be experimentally measured was found to be associated with expertise (van der Gijp et al., 2017).

Despite the development of this consensus feature set, visual search complexity may not be adequately captured by simple, low-dimensional features that do not fully describe how visual perception relates to skill. Machine learning is well-suited to provide deeper insight into radiologist visual search behavior and how this relates to radiologist performance. Waite et al. (Waite et al., 2019) highlighted the importance of understanding perceptual expertise in radiology and the potential use of eye-tracking and perceptual learning methods in medical training to improve diagnostic accuracy. Lim et al. (Lim et al., 2022) identified several features that can be extracted from eye-tracking data, including pupil size, saccade, fixations, velocity, blink, pupil position, electrooculogram (EOG), and gaze point to be used in Machine Learning models. Among these features, fixation was the most commonly used feature in the studies reviewed.

Shamyuktha and Amudha (S et al., 2022) developed a machine learning framework using eye gaze data such as saccade latency and amplitude to classify expert and non-expert radiologists. Harezlak et al. (Harezlak et al., 2018) investigated eye movement traits to differentiate experts and laymen in a similar study. Akshay et al. (Akshay et al., 2020) proposed a machine learning algorithm to identify eye movement metrics using raw eye-tracking data. Rizzo et al. (Rizzo et al., 2022) utilized machine learning to detect cognitive interference based on eye-tracking data. Öder et al.(ÖDER et al., 2022) applied machine learning to classify familiar web users based on eye-tracking data. Indeed, these techniques can be used to enhance competency assessment and feedback techniques in radiologists.

Eye-tracking also holds the potential for understanding the longitudinal aspects of competency progression in medical education, allowing for examining how interpretive and diagnostic skills develop over time. Karargyris et al. (Karargyris et al., 2021) and Bigolin Lanfredi et al. (Bigolin Lanfredi et al., 2022) created and validated chest X-ray datasets with eye-tracking data and report dictation for developing such A.I. systems. These datasets aim to support the research community in developing more complex support tools for radiology research.

In this study, we employ machine learning to compare the discriminability of two radiologists of different skill levels using, first, the aforementioned "traditional" gaze-based features (such as time to scan, saccade length, the total number of fixations, and total regressive fixations) (van der Gijp et al., 2017) and second, the "proposed" features that we developed to describe high dimensional visual search patterns thoroughly and quantitatively. We curate the traditional feature sets to those which could be practically acquired without laborious manual ground truthing of exams, as this would permit large-scale deployment of this technology to health care institutions. To highlight the use of eye-tracking data and artificial intelligence, we term our general approach "biometric radiology artificial intelligence."

The driving hypothesis behind the work presented in this paper is that gaze patterns measurably differ among radiologists as a function of their experience level. To test this hypothesis, we proposed a novel discretized feature encoding method that condenses fixation data into a few representative spatiotemporal bins for descriptive and predictive analytics purposes (See Figure 1).

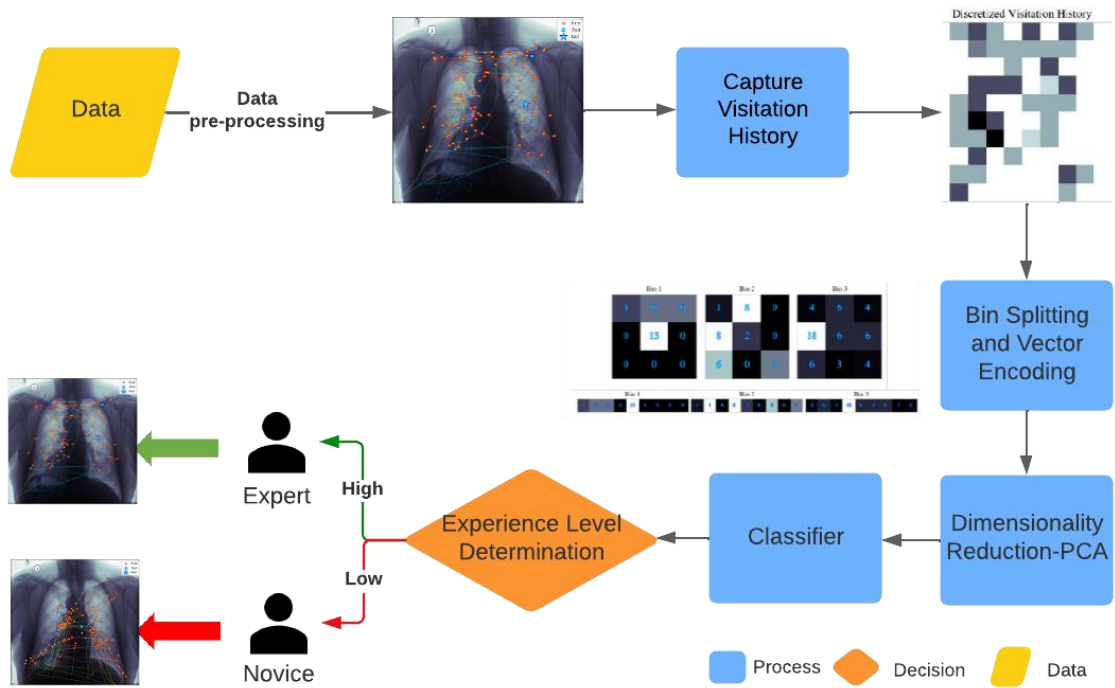

*Figure 1 - Overall Algorithm: The steps required to generate proposed features from the raw dataset and build the proposed machine learning model.*

Collecting data from two subjects, one faculty member (expert) and one resident (trainee), we analyze their behavior and level of experience using the proposed approach. Using stratified cross-validation over 10 folds, we compare the AUC performance of several classifiers using the proposed methodology with the AUC performance of those same classifiers when using a traditional feature set (see Table 1). We then confirm our results using data from a second similarly designed larger study evaluating 8 subjects: four faculty members (expert) and four residents (trainee). The remainder of the paper is structured as follows: Section 2 presents the data collection and preparation procedures and details of the proposed method. Section 3 describes the simulation study and interpretation. Finally, Section 4 presents the discussion, concluding remarks, and advice to practitioners.

## 2. Data Collection and Preparation:

The study design was prospective, controlled, block-randomized, and IRB approved. Each study subject completed four roughly one-hour sessions in a radiology reading room, including tutorial, calibration, assessment, and annotation periods.

The tutorial included an overview of the assessment period and instructions on how to perform dictation and annotation consistently. Calibration was performed to ensure that recorded and actual gaze were consistent based on a nine-point custom calibration mapping script.

Nodule and normal cases were derived from the Shiraishi 2000 chest radiograph dataset (Shiraishi et al., 2000), which includes 154 chest radiographs with 5 degrees of subtlety from level 1 (extremely subtle) to level 5 (obvious). Distractor cases were derived from the VinDr chest radiograph dataset (Nguyen et al., 2022). Three sets of six nodule cases from the JSRT dataset, one set each from the intermediate difficulty levels (2, 3, & 4), and one set of nine normal cases from the JSRT dataset were randomly sampled without replacement. Two cases each of pneumothorax, cardiomegaly, and consolidation from the VinDr dataset were randomly sampled without replacement to serve as distractor cases. These distractor cases functioned mainly to prevent control subject bias to the nodule detection task. No exam was reviewed twice by a study subject during the trial, and all study subjects reviewed the same set of cases.

A custom software tool was developed to automatically display the study images and capture timestamped bilateral gaze, bilateral pupil, head pose, voice, annotation, and image display configuration data. No chin rest was utilized to ensure that the study was performed in a manner that was as close as possible to a clinical setting. After each session, data was transferred to a database for further analysis.

## 2.1. Data Acquisition

In the first study, the EyeLink 1000 eye tracker and software were used to collect eye-tracking data (*EyeLink 1000 Plus*, n.d.). Two subjects, one faculty member (9 years of faculty experience) and one resident (3 years of trainee experience), scanned a series of chest x-ray images (CXR), which contained a balanced class composition of normal scans (no abnormalities), abnormal scans (mass/nodule present), or abnormal with pleural effusion. A total of 110 trials (55 trials per subject) were recorded. We leveraged the EyeLink suite to remove most artifacts, such as blinks, from the eye-tracking data captured in each subject's trial and manually filtered remaining artifacts, such as off-screen distractions left unprocessed (see the far-displaced fixations in Figure 2, for example).

In the second study, a Tobii 5L eye tracker was used (*Tobii Eye Tracker 5 | Next Generation of Head and Eye Tracking*, n.d.). This second dataset included 8 subjects (4 Faculty with an average of 12.75 years of faculty experience and 4 Trainees with 2.25 years of trainee experience), each scanning the same set of 27 images. The Tobii gaze data was unprocessed to evaluate the robustness of the proposed method to fixation post-processing.

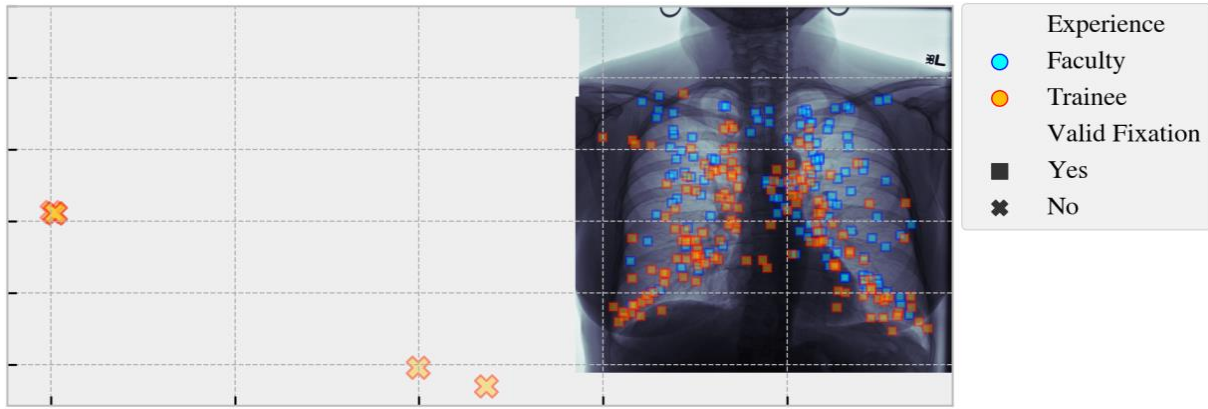

*Figure 2: Example of eye-tracking fixations for one trial processed by the EyeLink software. The fixations illustrated include subjects 1 (blue) and 2 (red) superimposed on the image displayed during the trial. The "invalid" fixations that were not successfully filtered out are shown as "x" markers and were manually removed during data processing.*

## *2.2. Common (Traditional) Features*

To establish a necessary baseline to which the proposed methodology can compare, several attributes were established based on a meta analysis done by Van der Gijp in 2017 (van der Gijp et al., 2017). We separated those features based on if they required ground truthing of exams. Ground truthing of medical exams are costly, time consuming, and error prone; this limits applicability and transferability to real-world applications. Consequently, features that required knowledge of the image abnormalities' ground truth location (i.e., area of interest) were removed: fixation duration on the area of interest, number of fixations on the area of interest, and the time between trial start and the first fixation on the area of interest. Table 1 summarizes the remaining attribute names, descriptions, and expected association with levels of expertise. All features were used as originally defined except for coverage. Salient regions refer to areas of an image that are not part of a peripheral black background. This is typically necessary because users may be viewing scans with different amounts of background area. As noted previously, we used the Tobii gaze data without fixation post-processing. For evaluating traditional features utilizing fixations in the Tobii dataset, we substituted raw gaze. For purposes of clarity and brevity, we use fixations and gaze interchangeably for the remainder of the paper.

*Table 1-State-of-the-art features from 22 relevant studies.*

| Attribute (Per trial) | Attribute Description | Association with High Level of Expertise (Percentage of the total number of included studies) |
|---|---|---|
| Total Time-to-Scan | Total amount of time taken to scan CXR image | Decrease (45.45%) |
| Regressive Fixation Count | The Total number of discretized CXR locations with more than 1 captured fixation | Increase (4.55%) or decrease (4.55%) |
| Fixation Count | Total captured fixations in a single CXR scan | Decrease (18.18%) |
| Total Saccade Length | Sum of saccade lengths (Robinson, 1997) (Time between fixations) captured in a single CXR scan | Increase (9.09%) or decrease (4.55%) |
| Coverage | Percent of salient regions in CXR images covered by scan | Increase (9.09%) or decrease (9.09%) |

*2.3. Proposed Approach: Discretized Vector Encoding for Fixation Data*

Here, we describe the proposed method for directly utilizing the fixation patterns as an alternative approach to using the current and previously described attributes in Table 1. The proposed strategy aims to extract information from fixations in two ways: (1) geometric alignment, such as the Cartesian locations of the fixations when displayed on a CXR image, and (2) temporal order in which the fixations appear. For each trial with recorded fixation data, we split the fixations into *t*-number of temporal bins or groups before counting the number of fixations captured within grids or subdivisions of size. Then, the *t*-number of grids is encoded into a single vector. The overall procedure is described in pseudocode in Algorithm 1 and illustrated in Figure 3.

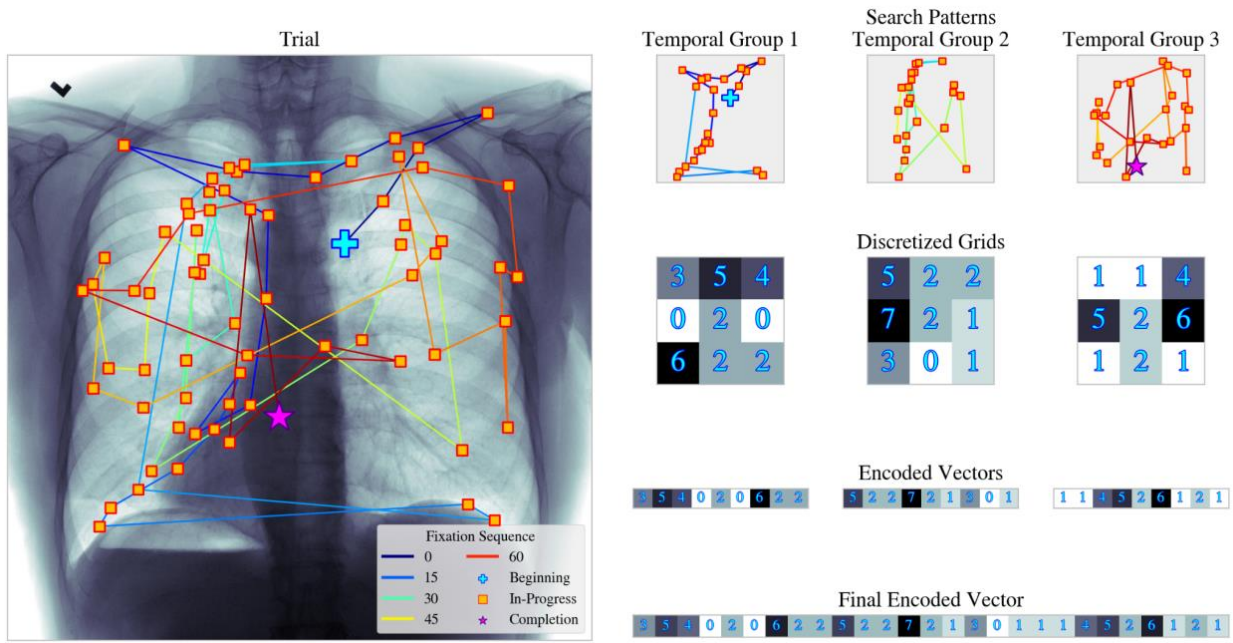

*Figure 3 - Proposed discretized vector encoding for fixation data. Bins 1, 2, & 3 capture fixations in a preserved spatial dimension across different temporal windows.*

In Figure 3, Subject 1 inspects a single chest X-ray image; the processed fixations are captured as illustrated gold/red squares, with the first fixation labeled as a blue cross and the last fixation visualized as a magenta star. In this example, the fixations are split into three temporal segments (Step 1), in which three-by-three grids count the number of fixations within them (Step 2). Then, the proposed algorithm outputs the final encoding vector as the flattening and concatenation of the set of three-by-three grids (Step 3). For a given (square) grid of size $x$ and $t$ number of temporal segments, the final yielded output vector is of length, regardless of trial temporal duration. Segmenting the raw data into fixed temporal segments is one of the benefits of this approach and a strategy developed and imposed to generate consistent numbers of variables on the encoding output across different trials. As the number of fixations across each trial can vary between subjects, fixing the number of temporal segments allows the capturing of trial duration while conforming to a prescribed number of grid subdivisions and temporal groups. For example, with Figure 3 as a reference, the final encoded vector will yield a vector with larger values therein for longer trials and yield a sparse vector (with lower values or values of zero) for shorter trials. Users can increase the fidelity of the grid and the number of time groups to represent a continuous spatiotemporal domain more closely. It is notable that the proposed methodology possesses the capability for tensor configuration for use in deep-learning architecture by using $t$ layers of **grids**. This tensor configuration is not studied in the paper due to the small sample size. The introduced technique is designed with more accessible or simpler classifiers in mind.

*Algorithm 1: Discrete Vector Encoding for Fixation Data*

| **Algorithm 1: Vector Encoding for Fixation Data** | |
|---|---|
| **Input** | $n$-Fixation coordinates of a single trial, $F_{[n\times 2]}$, number of x- and y-axis subdivisions, $(x, y)$, number of temporal groups, $(t)$ |
| **Output** | Encoded vector, $V_{[1\times(t*x*y)]}$ |
| **Initialize** | Create array $A_{[t\times x\times y]}$ and centroids $C_{[(x*y)\times 2]}$ corresponding to the center of each grid subdivision (defined by second and third indices of $A$ |
| | Evenly split fixations into $t$-groups, $T = ([F_1, F_2, \cdots]^1, \cdots, [\cdots, F_{(n-1)}F_n]^t)$ |
| **Procedure** | **For** $i = 1 \to t$ **do**: |
| | $\quad f = T_i$ |
| | $\quad$ **For** $j = 1 \to len(f)$ **do**: |
| | $\quad\quad C^* = argmin(\|\| C - f_j \|\|)$ |
| | $\quad\quad A[i, C_x^*, C_y^*] \mathrel{+}= 1$ |
| | $\quad V = vec(A)$ |
| | **Return** $V$ |

## 3. Analysis and Interpretation:

### *3.1 Competing Algorithms and Training*

In this study, we employ the *Gaussian process*, *Logistic regression*, and *K-nearest-neighbors* classifiers from the *Scikit-learn* (Pedregosa et al., n.d.) package, the *XG-Boost* (*XGBoost | Proceedings of the 22nd ACM SIGKDD International Conference on Knowledge Discovery and Data Mining*, n.d.) tree-based ensemble classifier, and a modified *AlexNet (Krizhevsky et al., 2012)* deep learning classifier. The Scikit-learn classifiers were selected for their accessibility to users, while the XG-Boost and AlexNet-Like neural networks were chosen as more complex classifiers. The Logistic regression, K-nearest neighbors, and XGBoost classifiers used Scikit-learn's *StratifiedKFold* and *GridSearchCV* packages to train on the *balanced accuracy* loss function (also defined by Scikit-learn), while the Gaussian process methodology used Laplace approximation as detailed in their documentation (Pedregosa et al., n.d.). Lastly, the AlexNet-like classifier employed sparse categorical cross-entropy (Cybenko et al., 1999) for training.

## 3.2 Performance Metric and Simulation Setup

To evaluate the discriminability of subjects using the proposed approach, we employ a stratified-fold cross-validation to calculate the AUC metric for several classification models, where each of the folds contains 5 trials from both levels of experience as the hold-out set. The study was performed on the data acquired by the EyeLink and Tobii equipment separately, and the following sections will contain an elaboration on their respective results.  We performed cross-validations on a full-factorial configuration of 5, 7, 10, and 15 square grid subdivisions and 3, 5, 10, and 20 temporal groupings and selected the settings for each classifier that yielded the best results.  In the presentation of these results, the average scores were calculated by computing the AUC metrics at the lowest level (*data acquisition method, classifier, data type, feature extraction method, grid-size, temporal-group, and cross-validation seed*) and averaged to the presented levels of granularity. Given the small sample size of 110 (EyeLink dataset) and 216 (Tobii dataset) trials, and high dimensionality in the chosen configurations (up to 4,500 encoded variables in our study), there are available pathways that we have employed to alleviate the effects of the curse of dimensionality present (Aggarwal et al., 2001), such as principal component analysis (PCA) (Wold et al., 1987) and kernel principal component analysis (KPCA) (Schölkopf et al., 1997). The feature extraction and dimensionality reduction methods employed include reducing the input data to two dimensions (with varying amounts of explained variance) and fixing the amount of variance explained to 50%, 90%, and 99% (with varying numbers of dimensions). These techniques were utilized not only to reduce the density of the data but also to introduce an additional pre-processing step that leverages the spectral decomposition of data collected from each subject.

Some of the major reasons for considering PCA and KPCA instead of the other alternatives include: (1) PCA and KPCA are among the most popular method of dimensionality reduction, (2) most technical practitioners, especially in the field of medicine, are familiar with PCA and KPCA, (3) PCA and KPCA have rigorous mathematical properties and commonly used baseline methods in statistical analysis, (5) PCA and KPCA have relatively low computational complexity compared to many of the other shallow and deep alternatives.

### 3.3 Results - EyeLink Dataset

Figure 4 illustrates the average AUC across each classifier tasked with distinguishing between two subjects (particularly between two levels of experience) using either the traditional or the proposed encoded data types (features). Along with the original data types, we include the average AUC of the classifiers based on the usage of select feature extraction configurations. The encoded features extracted from the raw dataset, shown on the left, illustrate a consistently high AUC score compared to the traditional features shown on the right, implying that the model performance for each classifier (except for certain feature extraction configurations of the AlexNet model) has high discriminatory power under optimal spatiotemporal encoding settings.

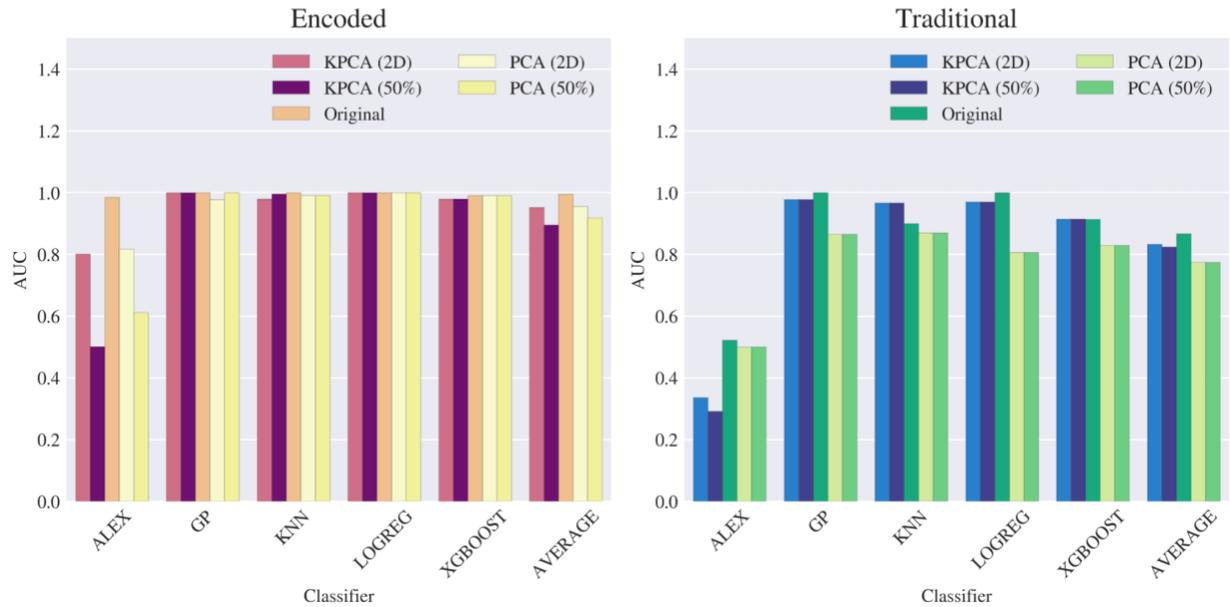

*Figure 4 - Numerical study results on the AUC metric reported for each classifier when consuming the EyeLink dataset, organized by the aggregated average of classifier, data type, and select feature extraction levels.*

We also present the AUC metrics from Figure 4 below in Table 2. The performance of the classifiers utilizing the encoded data type consistently yielded higher discriminatory power than those utilizing the traditional data type across all feature extraction methods. Encoding the fixation data into the proposed spatiotemporal elements provides more information each classifier can use to determine the experience level of a given subject more effectively than utilizing traditional attributes. This table illustrates the original encoded data to possess the highest performance, with AUC scores consistently above 0.98 across all classifiers. However, usage of the traditional data yields roughly 0.522 at worst, as seen in the reported results for the AlexNet classifier. This trend of encoded data providing better results is also seen when utilizing feature extraction; though a performance decrease is observable when reducing dimensions either through an information covariance matrix (PCA) or spatial relation (KPCA), the use of encoded data still outperforms those corresponding to the employment of the traditional data. This suggests that the loss in information due to dimensionality reduction can be considered negligible in light of the benefits of utilizing spatiotemporal encoding. The lower relative performance of the AlexNet-like classifier is likely caused by the number of training samples available in this study. The report on AUC in the table for the classifier is higher for the encoded data type, where it is observable that utilizing the data without dimensionality reduction provides the best performance. This effect has been studied in (Sumner & Alaeddini, 2019) in which neural networks already perform feature extraction throughout each present layer; this supportively evidences the reported results here, whereas (besides the small dataset) performing feature extraction beforehand may not provide enough information for the network to use its architecture to its fullest potential.

*Table 2 – Numerical Tabulation of AUC scores across each classifier and data type and select feature extraction methods.*

| **EyeLink Dataset - AUC** | | | | | | |
|---|---|---|---|---|---|---|
| **Feature Extraction** | **Data Type** | **Classifier** | | | | |
| | | **ALEX** | **GP** | **KNN** | **LOGREG** | **XGBOOST** |
| KPCA (2D) | Encoded | 0.801 | 1.000 | 0.980 | 1.000 | 0.980 |
| | Traditional | 0.336 | 0.978 | 0.967 | 0.970 | 0.914 |
| KPCA (50%) | Encoded | 0.501 | 1.000 | 0.996 | 1.000 | 0.980 |
| | Traditional | 0.292 | 0.978 | 0.967 | 0.970 | 0.914 |
| Original | Encoded | 0.985 | 1.000 | 1.000 | 1.000 | 0.991 |
| | Traditional | 0.522 | 1.000 | 0.900 | 1.000 | 0.914 |
| PCA (2D) | Encoded | 0.818 | 0.978 | 0.991 | 1.000 | 0.991 |
| | Traditional | 0.500 | 0.866 | 0.870 | 0.806 | 0.830 |
| PCA (50%) | Encoded | 0.611 | 1.000 | 0.991 | 1.000 | 0.991 |
| | Traditional | 0.500 | 0.866 | 0.870 | 0.806 | 0.830 |

By using the encoded vectors for classification, differences in eye-tracking patterns can more consistently be distinguished between the two subjects. Figure 5 shows one such difference in search pattern behavior. The more experienced subject (subject 1, 5a) shows a more uniformly distributed search pattern across the chest X-ray. In contrast, the less-experienced subject (subject 2, 5b) focuses on regions where they suspect abnormalities. It is clear from a visual perspective that the behavior between these subjects is markedly different, and using the correct spatiotemporal configurations to capture the differences between the two subject's behavior by leveraging the proposed methodology (as reported numerically in Table 2) provides a consistent improvement of classification accuracy.

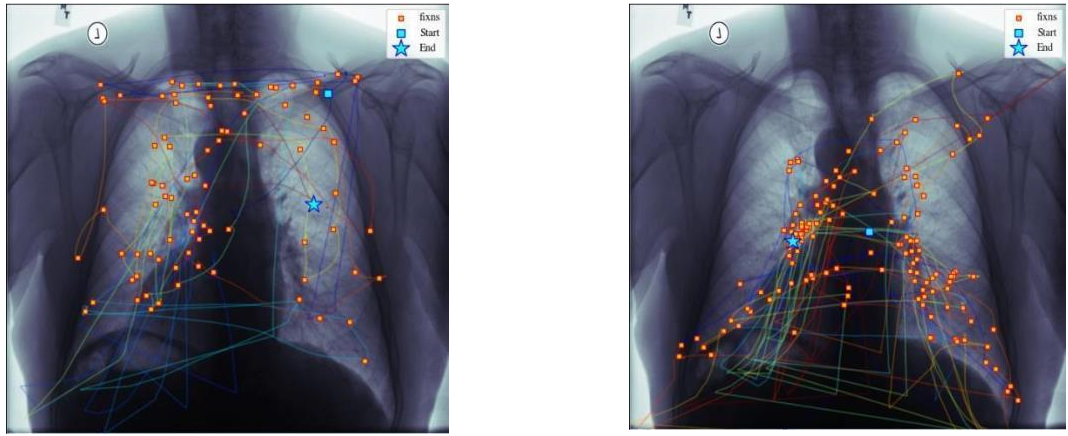

(a) Subject 1 (Faculty) Scan of Chest X-ray

(b) Subject 2 (Trainee) Scan of Chest X-ray

*Figure 5 – Search pattern comparison between the two studied subjects on an x-ray containing an abnormality. (a) Subject 1 performs a less-targeted search while Subject 2 (b) spends more focus on regions in which the subject believes abnormalities may be present.*

### *3.4 Results - Tobii Dataset*

We have performed the same analysis on the data acquired using Tobii eye-tracking equipment. It is notable that though the AUC scores from the EyeLink dataset are consistently high, natural anticipations include one to observe more variation in classifier performance when more individual subjects (classified as either a more-experienced Faculty or less-experienced Trainee) are introduced to the study. Figure 6 illustrates a report on AUC in a similar fashion to that in Figure 4, with lower scores across all classifying models for both data types. As seen in Figure 4, Figure 6 also suggests that the best performance for the encoded data on average is attained when utilizing it without feature extraction, though, for several cases, we can observe that some form of feature extraction yields better results than their respective traditional dataset counterparts.

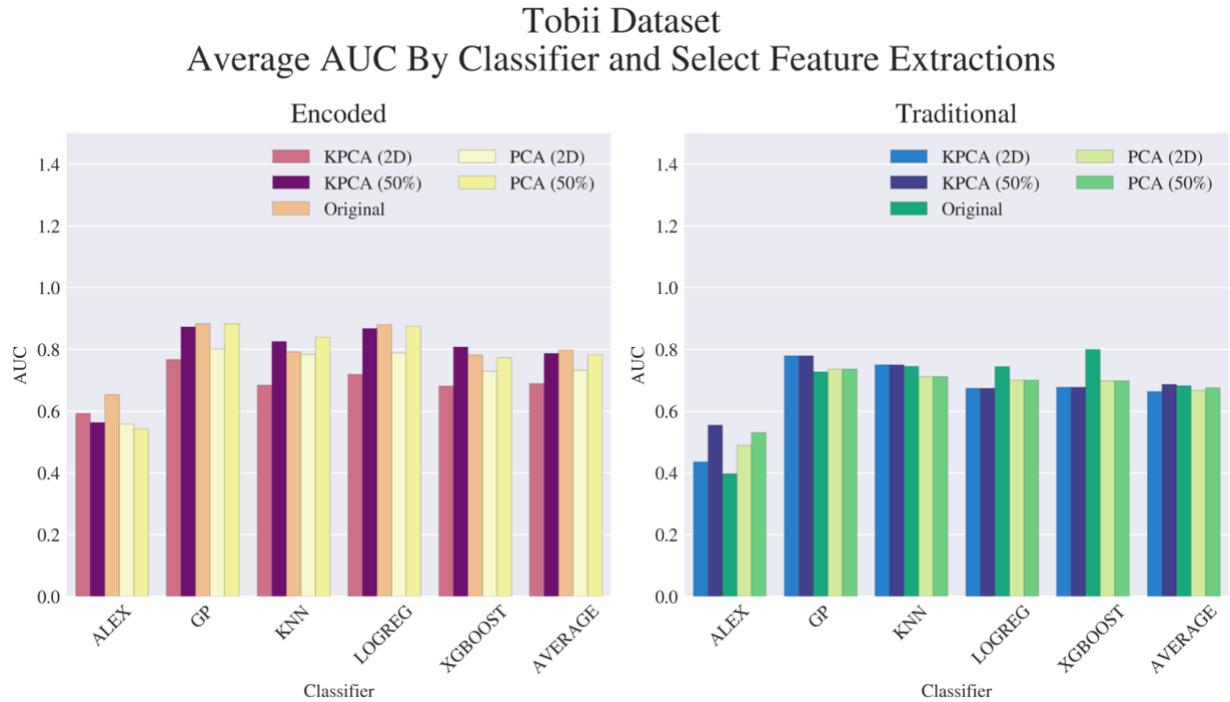

*Figure 6 - Numerical study results on the AUC metric reported for each classifier when consuming the Tobii dataset, organized by the aggregated average of classifier, data type, and select feature extraction levels.*

When inspecting Tables 3-7, we can numerically inspect the average (and variance of) AUC, F1-Score, Accuracy, Specificity, and Sensitivity of each classifier when consuming each data type in both datasets. Within the Tobii dataset, the encoded data type generally outperformed (shown in bold) the traditional data type across most metrics and models. Though the encoded data type that was consumed within the Tobii dataset possessed more discriminatory capability than did the traditional data, the performance gap was less pronounced than those observable in the EyeLink dataset. For example, the Tobii average (and variance) AUC scores for the encoded data type ranged from 0.55 (+/- 0.05) to 0.82 (+/- 0.04), while the traditional data type ranged from 0.51 (+/-0.07) to 0.76 (+/- 0.05), and the EyeLink average (variance) AUC scores for the same data types ranged from 0.63 (+/- 0.07) to 1.0 (+/- 0.0) and 0.52 (+/- 0.08) to 0.96 (+/- 0.01), respectively. There is a consistent trend across datasets that support the encoded data is capable of providing higher values of accuracy and performance; The F1 score for the Gaussian process, K-nearest-neighbors (KNN), Logistic regression (LogReg), and XGBoost were consistently higher when using the encoded data than when using traditional attributes in classification. This highlights the ability of the proposed encoding procedure to improve the balance between precision and recall in the classifiers and, as a result, the overall effectiveness of each model's predictions. In terms of specificity, the encoded data type is also shown to have a competitive edge in boosting a classifier's ability to correctly identify true negative class labels (experienced subjects). As seen in Table 6, the average range of improvement lies between 0.01 to 0.04 for the EyeLink dataset, and -0.2 to +0.15 for the Tobii dataset; the negative value of the improvement is seen with the AlexNet-like model, which, as explained before, may have difficulty fitting well for classification on small datasets, made more difficult by the variation in subconscious behavior between subjects that are recorded in spatiotemporal encodings by the proposed methodology.

*Table 3 – Numerical Tabulation of mean and variance AUC scores across each data acquisition method, classifier, and data type.*

| | | | Metric: AUC | | | |
|---|---|---|---|---|---|---|
| **Data Acquisition** | **Classifier** | **ALEX** | **GP** | **KNN** | **LOGREG** | **XGBOOST** |
| | **Data Type** | | | | | |
| **EyeLink** | **Encoded** | **0.63 (± 0.07)** | **1.0 (± 0.0)** | **0.96 (± 0.01)** | **1.0 (± 0.0)** | **0.97 (± 0.0)** |
| | Traditional | 0.52 (± 0.08) | 0.96 (± 0.01) | 0.93 (± 0.01) | 0.93 (± 0.01) | 0.91 (± 0.01) |
| **Tobii** | **Encoded** | **0.55 (± 0.05)** | **0.82 (± 0.04)** | 0.73 (± 0.05) | **0.82 (± 0.04)** | **0.73 (± 0.05)** |
| | Traditional | 0.51 (± 0.07) | 0.76 (± 0.05) | **0.74 (± 0.04)** | 0.71 (± 0.08) | 0.71 (± 0.05) |

*Table 4 – Numerical Tabulation of mean and variance F1 scores across each data acquisition method, classifier, and data type.*

| | | | Metric: F1 | | | |
|---|---|---|---|---|---|---|
| **Data Acquisition** | **Classifier** | **ALEX** | **GP** | **KNN** | **LOGREG** | **XGBOOST** |
| | **Data Type** | | | | | |
| **EyeLink** | **Encoded** | **0.43 (± 0.1)** | **0.98 (± 0.0)** | **0.9 (± 0.03)** | **0.98 (± 0.0)** | **0.96 (± 0.0)** |
| | Traditional | 0.41 (± 0.09) | 0.9 (± 0.03) | 0.88 (± 0.02) | 0.82 (± 0.04) | 0.87 (± 0.02) |
| **Tobii** | **Encoded** | **0.39 (± 0.1)** | **0.73 (± 0.07)** | 0.61 (± 0.1) | **0.74 (± 0.06)** | **0.68 (± 0.07)** |
| | Traditional | 0.23 (± 0.1) | 0.71 (± 0.05) | **0.71 (± 0.04)** | 0.58 (± 0.09) | 0.67 (± 0.05) |

*Table 5 – Numerical Tabulation of mean and variance Accuracy scores across each data acquisition method, classifier, and data type.*

| Metric: Accuracy | | | | | | |
|---|---|---|---|---|---|---|
| **Data Acquisition** | **Classifier** | **ALEX** | **GP** | **KNN** | **LOGREG** | **XGBOOST** |
| | **Data Type** | | | | | |
| **EyeLink** | **Encoded** | **0.52 (± 0.02)** | **0.99 (± 0.0)** | **0.93 (± 0.01)** | **0.98 (± 0.0)** | **0.96 (± 0.0)** |
| | Traditional | 0.49 (± 0.04) | 0.93 (± 0.01) | 0.9 (± 0.01) | 0.88 (± 0.01) | 0.9 (± 0.01) |
| **Tobii** | **Encoded** | **0.51 (± 0.0)** | **0.76 (± 0.03)** | 0.69 (± 0.03) | **0.77 (± 0.03)** | **0.72 (± 0.03)** |
| | Traditional | 0.5 (± 0.01) | 0.7 (± 0.03) | **0.7 (± 0.02)** | 0.64 (± 0.03) | 0.67 (± 0.03) |

*Table 6 – Numerical Tabulation of mean and variance Specificity scores across each data acquisition method, classifier, and data type.*

| Metric: Specificity | | | | | | |
|---|---|---|---|---|---|---|
| **Data Acquisition** | **Classifier** | **ALEX** | **GP** | **KNN** | **LOGREG** | **XGBOOST** |
| | **Data Type** | | | | | |
| **EyeLink** | **Encoded** | **0.43 (± 0.22)** | **0.98 (± 0.0)** | **0.97 (± 0.01)** | **0.99 (± 0.0)** | **0.95 (± 0.01)** |
| | Traditional | 0.42 (± 0.22) | 0.97 (± 0.01) | 0.93 (± 0.01) | 0.97 (± 0.01) | 0.92 (± 0.01) |
| **Tobii** | **Encoded** | 0.46 (± 0.16) | **0.8 (± 0.04)** | **0.76 (± 0.06)** | **0.81 (± 0.04)** | **0.76 (± 0.05)** |
| | Traditional | **0.66 (± 0.12)** | 0.65 (± 0.07) | 0.67 (± 0.04) | 0.67 (± 0.08) | 0.63 (± 0.08) |

*Table 7 – Numerical Tabulation of mean and variance Sensitivity scores across each data acquisition method, classifier, and data type.*

| Data Acquisition | Classifier / Data Type | ALEX | GP | KNN | LOGREG | XGBOOST |
|---|---|---|---|---|---|---|
| **EyeLink** | Encoded | **0.64 (± 0.22)** | **0.99 (± 0.0)** | **0.89 (± 0.05)** | **0.98 (± 0.01)** | **0.98 (± 0.0)** |
|  | Traditional | 0.57 (± 0.2) | 0.89 (± 0.04) | 0.87 (± 0.03) | 0.77 (± 0.06) | 0.86 (± 0.03) |
| **Tobii** | Encoded | **0.56 (± 0.17)** | 0.72 (± 0.07) | 0.61 (± 0.1) | **0.72 (± 0.06)** | 0.68 (± 0.07) |
|  | Traditional | 0.32 (± 0.12) | **0.76 (± 0.05)** | **0.74 (± 0.04)** | 0.6 (± 0.09) | **0.71 (± 0.05)** |

Metric: Sensitivity

## 4. Discussion and Conclusion

The majority of errors related to chest X-ray diagnoses have been attributed by literature to perceptual errors (Bruno et al., 2015). Due to the lack of a gold-standard search pattern for X-ray images, radiologists have developed unique search patterns over time during practice and under guidance from more experienced clinicians. Several authors have investigated the use of eye-tracking to reveal search patterns. Some studies provide fixation-based features that correlate to the level of experience, (Bernal et al., 2014) utilizing visual gaze to provide statistical differences between novices and experts. Similarly, laymen and experts were employed (Harezlak et al., 2018) to scan X-ray images, and the resulting eye-tracking data was used in conjunction with inferential statistics to classify the test subjects into the appropriate level of experience. These studies evidenced the capability to discriminate between subjects using eye-tracking information and provided a baseline concept that aided in the development of the research presented in this article.

Spatiotemporal features were generated from eye-tracking data using the proposed strategy; utilizing the encoding approach allows classifiers to better distinguish between subjects. The encoded vectors can capture both spatial and temporal information, which can vary greatly between subjects, as shown in Figure 5. Using several classifiers in the outlined study, the encoding approach provides higher measures of class separability (AUC) than utilizing common eye-tracking metrics as features. The results highlight the usefulness of the proposed encoding technique over utilizing traditional features across a range of metrics and models, regardless of which equipment was used to capture the data (EyeLink or Tobii). Though the number of samples is small across the two sets of data, the improvement despite the variation in behavior as a result of using 8 different subjects in one dataset emphasizes the robustness of the encoded methodology and its capability to transform captured eye-tracking data into attributes of interconnected spatial and chronological components. The results above suggest that the benefits of encoding eye-tracking data can also be generalized across a variety of applications involving user habits when scanning or observing an item of interest. Other than the small sample size, and while the results show a promising proof-of-concept from which more research can be conducted, there were other notable limitations to the study. Firstly, the difference in the data acquisition types used for analysis; the Tobii dataset did not utilize post-processed gaze features such as fixations and saccades. The traditional feature sets between the EyeLink and Tobii datasets are not identical, and the encoded attributes yield larger values in each output vector (though this difference did not greatly impact the trends between the two datasets in the presented results). In addition, metrics pertaining to the location of the abnormality were not included as features in the traditional data type during the performance evaluation (e.g., time to first fixation on the region of abnormality). Eye-tracking metrics involving the ground truth location have previously demonstrated statistical significance. Establishing ground truth, however, requires extensive labeling, validation, and additional processing that significantly limits the potential for deployment to educational and clinical settings. Secondly, the studied chest X-rays only represented three categories: normal, pleural effusion, and pulmonary nodule. It is possible that the performance of the feature encoding method may change when a full spectrum of X-ray abnormalities is evaluated.

Since developing a unique and accurate search pattern requires time and practice, we proposed and investigated the effects of using an alternative method based on (spatiotemporal) feature extraction to discern the identity of two radiologist subjects with distinct levels of expertise. This method aims not only to discriminate between subjects but creates a pathway to better characterize and understand trainee and faculty techniques. Through this improved understanding, it is possible to craft more powerful educational programs which can optimize search patterns. A good search pattern should quickly and efficiently cover areas of interest with high accuracy. With the implementation of this method as a source of knowledge transfer, we believe that the prevalence of radiological perceptual errors on normal chest X-rays may decrease, helping to preserve patients' lives and giving them a chance to receive the correct treatment for a given diagnosis.

In summary, we have shown the potential for spatiotemporal features extracted from eye-tracking data to be useful in discriminating between radiologists of different skill levels and opening the door to improving education. We plan to augment this research by increasing the number of radiologists to demonstrate generalizability and exploring additional types of spatiotemporal analyses.

**Image Source:**

Images and annotations were obtained from the NIH (National Institutes of Health) Clinical Center and can be downloaded at (Nguyen et al., 2022) and the Japanese Society of Radiological Technology (Shiraishi et al., 2000)

**Declarations** :

**Ethics approval and consent to participate:** The Ethics approval and consent to participate is provided in the IRB 19-533H. The study was performed under a protocol approved by the University of Texas, San Antonio Institutional Review Board, and participants gave written informed consent before testing.

**Consent for publication:** Not applicable.

**Availability of data and materials:** Images and annotations were obtained from the NIH Clinical Center and can be downloaded at https://nihcc.app.box.com/v/ChestXray-NIHCC (Nguyen et al., 2022) and the Japanese Society of Radiological Technology (Shiraishi et al., 2000). The IRB restricts sharing of the gaze data. In case of further information, please contact Kal L. Clark (clarkkl@uthscsa.edu).

**Competing interests:** There are no competing interests to declare.

**Funding:** This research is funded by San Antonio Medical Foundation (SAMF), PI: Kal Clark (UTHSA), and Ed Golob (UTSA).

**Authors' contributions:**

Data collection and medical knowledge was provided by KC, AA, and SE. Psychological analysis and perception-related error research were done by JM and EG. Engineering features were created by SM, and Data Analysis, methodology, experiments, and draft manuscript were done by SM, CR, SHAF, AA, and NC. All authors read and approved the final manuscript.

**References**

Aggarwal, C. C., Hinneburg, A., & Keim, D. A. (2001). On the Surprising Behavior of Distance Metrics in High Dimensional Space. In J. Van den Bussche & V. Vianu (Eds.), *Database Theory—ICDT 2001* (pp. 420–434). Springer. https://doi.org/10.1007/3-540-44503-X_27

Akshay, S., Megha, Y. J., & Shetty, C. B. (2020). Machine Learning Algorithm to Identify Eye Movement Metrics using Raw Eye Tracking Data. *2020 Third International Conference on Smart Systems and Inventive Technology (ICSSIT)*, 949–955. https://doi.org/10.1109/ICSSIT48917.2020.9214290

*An Update on Cancer Deaths in the United States*. (2022, February 28). Centers for Disease Control and Prevention. https://www.cdc.gov/cancer/dcpc/research/update-on-cancer-deaths/index.htm

Bernal, J., Sánchez, F. J., Vilariño, F., Arnold, M., Ghosh, A., & Lacey, G. (2014). Experts vs. novices: Applying eye-tracking methodologies in colonoscopy video screening for polyp search. *Proceedings of the Symposium on Eye Tracking Research and Applications*, 223–226. https://doi.org/10.1145/2578153.2578189

Bigolin Lanfredi, R., Zhang, M., Auffermann, W. F., Chan, J., Duong, P.-A. T., Srikumar, V., Drew, T., Schroeder, J. D., & Tasdizen, T. (2022). REFLACX, a dataset of reports and eye-tracking data for localization of abnormalities in chest x-rays. *Scientific Data*, *9*(1), Article 1. https://doi.org/10.1038/s41597-022-01441-z

Bruno, M., Walker, E., & Abujudeh, H. (2015). Understanding and Confronting Our Mistakes: The Epidemiology of Error in Radiology and Strategies for Error Reduction. *RadioGraphics*, *35*, 1668–1676. https://doi.org/10.1148/rg.2015150023

Cybenko, G., O'Leary, D. P., & Rissanen, J. (Eds.). (1999). *The Mathematics of Information Coding, Extraction and Distribution* (Vol. 107). Springer. https://doi.org/10.1007/978-1-4612-1524-0

Drew, T., Evans, K., Võ, M. L.-H., Jacobson, F. L., & Wolfe, J. M. (2013). Informatics in Radiology: What Can You See in a Single Glance and How Might This Guide Visual Search in Medical Images? *RadioGraphics*, *33*(1), 263–274. https://doi.org/10.1148/rg.331125023

*EyeLink 1000 Plus*. (n.d.). Fast, Accurate, Reliable Eye Tracking. Retrieved May 11, 2023, from https://www.sr-research.com/eyelink-1000-plus/

Felson's Principles of Chest Roentgenology: A Programmed Text. 2nd ed. (2000). *Radiology*, *214*(3), 848–848. https://doi.org/10.1148/radiology.214.3.r00fe55848

Harezlak, K., Kasprowski, P., & Kasprowska, S. (2018). Eye Movement Traits in Differentiating Experts and Laymen. In A. Gruca, T. Czachórski, K. Harezlak, S. Kozielski, & A. Piotrowska (Eds.), *Man-Machine Interactions 5* (pp. 82–91). Springer International Publishing. https://doi.org/10.1007/978-3-319-67792-7_9

Jones, M. (2014, January 8). Screening for Lung Cancer: U.S. Preventive Services Task Force Recommendation Statement. *Lung Cancer Research Foundation*. https://www.lungcancerresearchfoundation.org/screening-for-lung-cancer-u-s-preventive-services-task-force-recommendation-statement/

Karargyris, A., Kashyap, S., Lourentzou, I., Wu, J. T., Sharma, A., Tong, M., Abedin, S., Beymer, D., Mukherjee, V., Krupinski, E. A., & Moradi, M. (2021). Creation and validation of a chest X-ray dataset with eye-tracking and report dictation for AI development. *Scientific Data*, *8*(1), Article 1. https://doi.org/10.1038/s41597-021-00863-5

Kelly, B. S., Rainford, L. A., Darcy, S. P., Kavanagh, E. C., & Toomey, R. J. (2016). The Development of Expertise in Radiology: In Chest Radiograph Interpretation, "Expert" Search Pattern May Predate "Expert" Levels of Diagnostic Accuracy for Pneumothorax Identification. *Radiology*, *280*(1), 252–260. https://doi.org/10.1148/radiol.2016150409

Krizhevsky, A., Sutskever, I., & Hinton, G. E. (2012). ImageNet Classification with Deep Convolutional Neural Networks. *Advances in Neural Information Processing Systems*, *25*. https://proceedings.neurips.cc/paper_files/paper/2012/hash/c399862d3b9d6b76c8436e924a68c45b-Abstract.html

Kundel, H. L. (1989). Perception Errors in Chest Radiography. *Seminars in Respiratory Medicine*, *10*(3), 203–210. https://doi.org/10.1055/s-2007-1006173

Kundel, H. L., & Nodine, C. F. (1975). Interpreting Chest Radiographs without Visual Search. *Radiology*, *116*(3), 527–532. https://doi.org/10.1148/116.3.527

Lim, J. Z., Mountstephens, J., & Teo, J. (2022). Eye-Tracking Feature Extraction for Biometric Machine Learning. *Frontiers in Neurorobotics*, *15*. https://www.frontiersin.org/articles/10.3389/fnbot.2021.796895

Nguyen, H. Q., Lam, K., Le, L. T., Pham, H. H., Tran, D. Q., Nguyen, D. B., Le, D. D., Pham, C. M., Tong, H. T. T., Dinh, D. H., Do, C. D., Doan, L. T., Nguyen, C. N., Nguyen, B. T.,

Nguyen, Q. V., Hoang, A. D., Phan, H. N., Nguyen, A. T., Ho, P. H., … Vu, V. (2022). VinDr-CXR: An open dataset of chest X-rays with radiologist's annotations. *Scientific Data*, *9*(1), Article 1. https://doi.org/10.1038/s41597-022-01498-w

ÖDER, M., ERASLAN, Ş., & YESİLADA, Y. (2022). Automatically classifying familiar web users from eye-tracking data:a machine learning approach. *Turkish Journal of Electrical Engineering and Computer Sciences*, *30*(1), 233–248. https://doi.org/10.3906/elk-2103-6

Pedregosa, F., Varoquaux, G., Gramfort, A., Michel, V., Thirion, B., Grisel, O., Blondel, M., Prettenhofer, P., Weiss, R., Dubourg, V., Vanderplas, J., Passos, A., & Cournapeau, D. (n.d.). Scikit-learn: Machine Learning in Python. *MACHINE LEARNING IN PYTHON*.

Rizzo, A., Ermini, S., Zanca, D., Bernabini, D., & Rossi, A. (2022). A Machine Learning Approach for Detecting Cognitive Interference Based on Eye-Tracking Data. *Frontiers in Human Neuroscience*, *16*. https://www.frontiersin.org/articles/10.3389/fnhum.2022.806330

Robinson, K. A. (1997). Dictionary of Eye Terminology. *British Journal of Ophthalmology*, *81*(11), 1021–1021. https://doi.org/10.1136/bjo.81.11.1021c

S, S. R., J, A., & K, A. M. (2022). A Machine Learning Framework for Classification of Expert and Non-Experts Radiologists using Eye Gaze Data. *2022 IEEE 7th International Conference on Recent Advances and Innovations in Engineering (ICRAIE)*, *7*, 314–320. https://doi.org/10.1109/ICRAIE56454.2022.10054277

Schölkopf, B., Smola, A., & Müller, K.-R. (1997). Kernel principal component analysis. In W. Gerstner, A. Germond, M. Hasler, & J.-D. Nicoud (Eds.), *Artificial Neural Networks—ICANN'97* (pp. 583–588). Springer. https://doi.org/10.1007/BFb0020217

Shiraishi, J., Katsuragawa, S., Ikezoe, J., Matsumoto, T., Kobayashi, T., Komatsu, K., Matsui, M., Fujita, H., Kodera, Y., & Doi, K. (2000). Development of a digital image database for chest radiographs with and without a lung nodule: Receiver operating characteristic analysis of radiologists' detection of pulmonary nodules. *AJR. American Journal of Roentgenology*, *174*(1), Article 1. https://doi.org/10.2214/ajr.174.1.1740071

Sumner, J., & Alaeddini, A. (2019). Analysis of Feature Extraction Methods for Prediction of 30-Day Hospital Readmissions. *Methods of Information in Medicine*, *58*(6), 213–221. https://doi.org/10.1055/s-0040-1702159

Swensson, R. G. (1980). A two-stage detection model applied to skilled visual search by radiologists. *Perception & Psychophysics*, *27*(1), 11–16. https://doi.org/10.3758/BF03199899

Tanoubi, I., Tourangeau, M., Sodoké, K., Perron, R., Drolet, P., Bélanger, M.-È., Morris, J., Ranger, C., Paradis, M.-R., Robitaille, A., & Georgescu, M. (2021). Comparing the Visual Perception According to the Performance Using the Eye-Tracking Technology in High-Fidelity Simulation Settings. *Behavioral Sciences*, *11*(3), 31. https://doi.org/10.3390/bs11030031

*Tobii Eye Tracker 5 | Next Generation of Head and Eye Tracking*. (n.d.). Tobii Gaming. Retrieved May 11, 2023, from https://gaming.tobii.com/product/eye-tracker-5/

van der Gijp, A., Ravesloot, C. J., Jarodzka, H., van der Schaaf, M. F., van der Schaaf, I. C., van Schaik, J. P. J., & Ten Cate, T. J. (2017). How visual search relates to visual diagnostic performance: A narrative systematic review of eye-tracking research in radiology. *Advances in Health Sciences Education: Theory and Practice*, *22*(3), Article 3. https://doi.org/10.1007/s10459-016-9698-1

Waite, S., Grigorian, A., Alexander, R. G., Macknik, S. L., Carrasco, M., Heeger, D. J., & Martinez-Conde, S. (2019). Analysis of Perceptual Expertise in Radiology – Current Knowledge and a New Perspective. *Frontiers in Human Neuroscience*, *13*, 213. https://doi.org/10.3389/fnhum.2019.00213

Wold, S., Esbensen, K., & Geladi, P. (1987). Principal component analysis. *Chemometrics and Intelligent Laboratory Systems*, *2*(1), 37–52. https://doi.org/10.1016/0169-7439(87)80084-9

Wolfe, J. M., Lyu, W., Dong, J., & Wu, C.-C. (2022). What eye tracking can tell us about how radiologists use automated breast ultrasound. *Journal of Medical Imaging*, *9*(04). https://doi.org/10.1117/1.JMI.9.4.045502

Wu, C.-C., & Wolfe, J. M. (2019). Eye Movements in Medical Image Perception: A Selective Review of Past, Present and Future. *Vision*, *3*(2), Article 2. https://doi.org/10.3390/vision3020032

*XGBoost | Proceedings of the 22nd ACM SIGKDD International Conference on Knowledge Discovery and Data Mining*. (n.d.). Retrieved May 11, 2023, from https://dl.acm.org/doi/10.1145/2939672.2939785